\documentclass{article}

\usepackage{arxiv}
\usepackage{graphicx}
\usepackage{amsmath}
\usepackage[utf8]{inputenc} 
\usepackage[T1]{fontenc}    
\usepackage{hyperref}       
\usepackage{url}            
\usepackage{booktabs}       
\usepackage{amsfonts}       
\usepackage{nicefrac}       
\usepackage{microtype}      
\usepackage{lipsum}

\title{Complementary reinforcement learning towards explainable agents}

\author{
 Jung Hoon Lee\\
  Allen Institute for Brain Science\\
  Seattle, WA 98109 \\
  \texttt{jungl@alleninstitute.org} \\
}

\begin{document}
\maketitle

\begin{abstract}
Reinforcement learning (RL) algorithms allow agents to learn skills and strategies to perform complex tasks without detailed instructions or expensive labelled training examples. That is, RL agents can learn, as we learn. Given the importance of learning in our intelligence, RL has been thought to be one of the key components to general artificial intelligence, and recent breakthroughs in deep reinforcement learning suggest that neural networks (NN) are natural platforms for RL agents. However, despite the efficiency and versatility of NN-based RL agents, their decision-making remains incomprehensible, reducing their utilities. To deploy RL into a wider range of applications, it is imperative to develop explainable NN-based RL agents. Here, we propose a method to derive a secondary comprehensible agent from a NN-based RL agent, whose decision-makings are based on simple rules. Our empirical evaluation of this secondary agent's performance supports the possibility of building a comprehensible and transparent agent using a NN-based RL agent. 
\end{abstract}

\section{Introduction}
Reinforcement learning (RL), inspired by our brain's reward-based learning, allows artificial agents to learn a wide range of tasks without detailed instructions or labeled training sets which are necessary for supervised learning \cite{Hertz1991,Sutton2017}. Given that RL agents' learning resembles our process of learning and that learning is essential to our intelligence, it seems natural to assume that RL is one of the key components to brain-like intelligent agents or general artificial intelligence. Recent breakthroughs \cite{Mnih2015,Silver2016} from DeepMind team showed that RL could train neural network (NN)-based agents to outperform humans in video-games and even 'Go', reigniting interests in RL and its applications in NN-based agents, and noticeable developments in NN-based RL agents have been reported since then; see \cite{Mirowski2017, Mnih2016, Oh2016,Plappert2018, Wang2016} for examples. 

However, despite rapid improvements in RL, NN-based RL agents' decision-making process remains incomprehensible to us. For instance, the AlphaGo's strategies employed during the match with Sedol Lee exhibited efficiency leading to victories, but the exact reasons behind its moves are unknown. AlphaGo demonstrated that incomprehensible decision-making can still be effective, but its effectiveness does not mean that it cannot be faulty. In ‘high stake problems’ \cite{Rudin2018}, any mistakes can be critical and need to be avoided. A loss in one Go match out of 100 matches is insignificant, but crashing a car once out of 100 drives is perilous and unacceptable.  If we comprehend the exact internal mechanisms of RL agents, we can correct their mistakes without negative impact on their performance. That is, ’transparent’ agents with comprehensible internal decision-making processes \cite{Lipton2016} are necessary to safely deploy RL agents into high stake problems.

Two earlier studies \cite{Hayes2017, Waa2017} showed that the decision-making processes of RL agents can be translated into human-readable descriptions. Their proposed algorithms, which fall into the 'post-hoc' interpretation approach \cite{Rudin2018}, do provide some insights into RL agents’ decision-making process, but they do not address how we correct (or fix) RL agents' actions. Then, how do we build transparent RL agents? We propose a secondary agent as a potential solution, which utilizes simple rules to choose the best action. This proposal is based on two ideas. First, the secondary agent's action can be analyzed because it utilizes simple rules. Second, the secondary agent can perform general tasks, if it takes advantage of trained RL agents. 

In this study, we propose a quasi-symbolic (QS) agent as a secondary agent and compare its performance to RL agents' performance. Specifically, QS agents learn, from RL agents’ behaviors, the values of transitions of states. After learning the values of transitions, QS agents identify the most valuable state-transitions (‘hub states’) and search for a sequence of actions (action plans) to reach one of the hub states. If QS agent cannot find a proper action plan to reach a hub state, they choose the best action by comparing the values of immediate transitions. In this study, we tested QS agents' performance using the 'lunar-lander' benchmark problem available in 'OpenAI Gym' environments \cite{Brockman2016}. Our results show that QS agents' performance is comparable to RL agents' performance. While our experiments are conducted in a simple environment, the results indicate that comprehensible agents with transparent decision-making process can be derived from RL agents.

\section{Related work}
While  deep learning (DL) spreads its influence, it remains unclear whether DL can safely be applied to high stake decision problems (e.g., medicine and criminal justice) \cite{Rudin2018}. DL agents rely on cascades of nonlinear interactions among computing nodes (i.e., neurons), but the interactions are too complex to be analyzed, which limits our mechanistic understanding of DL agents' decision-making. This means that despite extensive testings we cannot fully predict their actions. Thus, deploying DL agents to high stake problems may lead to critical failures; such failures have already been reported \cite{Rudin2018}. 

To deploy DL agents into high stake decisions, it is imperative to understand the exact process of their decisions. If reasoning behind their decisions become clear, developers could improve DL agents’ reliability, policy-makers could introduce regulations to ensure fairness of DL agents, and users could use DL agents more effectively \cite{Preece2018}. While we note the diverse aspects of DL explainability  \cite{Preece2018}, Lipton \cite{Lipton2016} pointed out that explainable intelligent agents can be obtained by analyzing the internal mechanisms or the agents’ behaviors, which were referred to as transparency and post-hoc interpretability, respectively. If we could understand the internal mechanisms, we will be able to fully understand DL agents' decision-making process; that is, their decisions will become 'transparent'. Several methods have been proposed to analyze the internal mechanisms of DL agents \cite{Zeiler2014,Yosinski2015,Fong2017,Erhan2009,olah2017feature}, and  Olah et al.\cite{olah2018the} presented an intriguing way to synergistically use them to gain insights into DL agents' decision-making. Also, multiple studies sought the post-hoc interpretability of DL agents, which we could use to predict and prevent potential failures. Specifically, human interpretable descriptions can be automatically generated by secondary agents \cite{McAuley2015,Lipton2016}, and representative examples or image parts can be identified by algorithms such as sensitivity analysis \cite{Samek2017,Montavon2018}. 

The majority of studies have focused on feedforward DL, but a few studies pursued the explainability of RL. Specifically, Hayes and Shah \cite{Hayes2017} showed that a secondary network could be trained to provide user-interpretable descriptions, and Waa et al. \cite{Waa2017} further showed that user-interpretable descriptions could be contrasive; that is, their methods can explain why RL agents would prefer one option to another \cite{Miller2019,Mittelstadt2019}. Although such post-hoc interpretability can be used to evaluate the quality/reliability of RL agents, it does not provide a way to fix RL agents' mistakes. However, with transparent agents, we can correct their mistakes selectively. In our study, we propose a potential approach, which can help us obtain transparent RL agents. 

\section{Network Structure}\label{network structure}
QS agents interact with RL agents and make plans for future actions by utilizing the model of environment (Env network). All these three networks/agents are constructed using the 'Pytorch', an open-source machine learning toolkit \cite{Paszke2017}. Below, we discuss QS, RL and Env network in details. 

\subsection{The structure of reference RL agent }
We used an actor-critic model as a reference RL agent \cite{Grondman2012, Konda2000}. The implementation was adopted from the official pytorch github repository \cite{Pytorch-team2018}, which maximizes the expected discounted reward $E_{\pi_{\theta}}[R]$, where $R=\sum^T_{t=0} \gamma^{t}r_t$, given policy $\pi_{\theta}$; where $T$ and $\theta$ represent a trajectory and parameters of the policy.  The gradient ascent of $E$ is estimated as shown in eq. (\ref{eq_grad}).

\begin{equation} \label{eq_grad}
\begin{aligned}
& \bigtriangledown_{\theta}[R]=E \big[ \sum^T_{t=0} \bigtriangledown_{\theta} \log \pi_{\theta} (a|s) \{r'(t)-V(s,t)\} \big] \\
& r'(t)=\frac{R(t)-<R>}{\sigma_R}
\end{aligned}
\end{equation}

, where $a$, $\theta$, $s$, $V=V(s,t)$ represent action, parameters, state and value function, respectively; where $<R>$ and $\sigma_R$ represent the mean and standard deviation of R in episodes. 

The actor consists of 8 input, 100 hidden and 4 output nodes, and the critic, 8 input, 100 hidden and 1 output nodes. The initial learning rate is 0.003 and is decreased by 10 times at every 1000 episode.

\begin{figure}[ht]
	\vskip 0.2in
	\begin{center}
		\centerline{\includegraphics[width=\columnwidth]{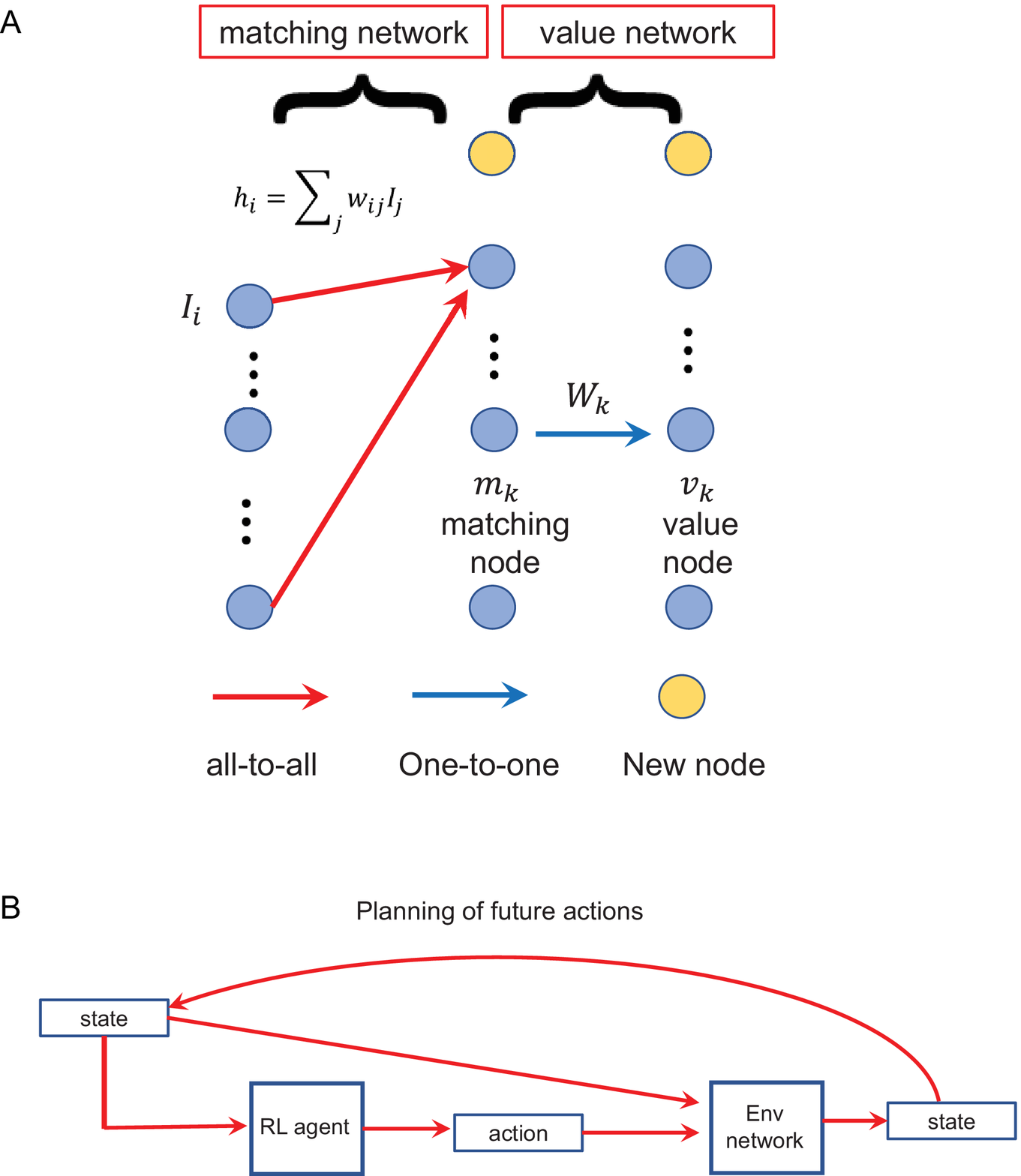}}
		\caption{The schematics of models. (A), the structure of the QS agents consisting of matching and value networks. The matching nodes (i.e., the outputs of matching network) are connected to the value nodes via one-to-one connection. When the input vector is novel (Eq. \ref{eq1}), new nodes are added to the matching and value networks and are connected via exclusive one-to-one connection. That is, the number of value nodes is the same as that of matching nodes. (B), the planning of future actions. At each time step, a QS agent invokes a RL agent to get a probable action and feeds it into Env network to predict the next state. By recursively using them, QS agents can make action plans.}
		\label{Fig1}
	\end{center}
	\vskip -0.2in
\end{figure}

\subsection{The structure of quasi-symbolic (QS) agent}
QS agent consists of matching and value networks, which are both single (synaptic) layer networks. The matching and value networks are sequentially connected (Fig. \ref{Fig1}A). That is, the output node of matching network is directly connected to the output node of value network. The number of matching nodes (output nodes in the matching network) is identical to the number of value nodes (output nodes in the value network), and the connections between matching and value networks are one-to-one. The matching network memorizes input vectors by imprinting normalized inputs to synaptic weights converging onto the output node $O_i$, as shown in Eq. \ref{eq1}; $h_i$ the synaptic input to $O_i$.

\begin{equation}\label{eq1}
h^k_{i}=\sum_{j}w_{ij}\frac{s^k_j}{\left\Vert\vec{s_k}\right\Vert^2}, \text{where } w_{mn}=\frac{s^m_n}{\left\Vert\vec{s_m}\right\Vert^2}
\end{equation}                               

, where superscript indices represent the inputs (state-transition vectors in this study; see below), and subscript indices represent the components of inputs. With normalized inputs stored in synaptic weights $w_{ij}$, the outputs of matching nodes represent cosine similarities between the current input and stored inputs. If all synaptic inputs $h_i$ to all output nodes $O_i$ are smaller than the threshold value $\theta $, the input is considered novel and stored into the matching network by adding a new output node into the matching network; the default threshold is $\theta=0.97$, unless stated otherwise. Once a new node is added to the matching network, a new node is also added to the value network, to keep one-to-one mapping (Fig. \ref{Fig1}A). The strength of the connection between these two newly established units is determined by the reward obtained by RL agents with the selected action. When the current input is one of the previously stored ones (i.e., one of synaptic input to output nodes is higher than the threshold $\theta $), the maximally activated node is identified, and the connection to the corresponding value network node is updated by adding the reward induced by the input vector (Eq. \ref{eq2}). It should be noted that only one matching node, which receives maximum synaptic input $h$, is allowed to be active one at a time, which makes the matching network operate in linear regimes.  

\begin{equation} \label{eq2}
W_k \leftarrow W_k+r_t
\end{equation}
, where $W_k$ is the connection between the matching node $m_k$ and value node $v_k$, and $r_t$ is the reward the RL agent obtain at time $t$. 

In this study, the matching network is designed to memorize the transitions between states $S$ and $S'$. That is, the input to the network is $\Delta S=S'-S$. Since there are 8 state variables in the lunar-lander benchmark problem, the number of input nodes of the matching network is 8. The sizes of the matching and value networks are not fixed. Instead, they are determined during the training of QS agents. The more diverse inputs are, the bigger matching networks become. For instance, if all inputs are identical, there is only one matching node. 

\subsection{Env network}

The Env network models the environment. Thus, it receives the state vector $S$ and action $a$ as inputs and returns the next state (Fig. \ref{Fig1}B). The inputs layer includes 12 nodes which represent 8 state variables and 4 possible actions in the lunar lander environment, and the output layers include 8 nodes (due to 8 state variables). In our experiments, we set the hidden layer of the Env network to have 300 nodes. The Env network is trained using the mean squared error (squared L2 norm). In each episode of RL training, the error is accumulated. After each episode, the Env network is updated using the accumulated error; that is, 1 episode is a single batch of backpropagation. The initial learning rate is 0.05 and is decreased to 10 percent at every 1000 episode. 

\section{Results}

In this section, we discuss the operating principles of QS agents including the interplay between QS and RL agents. Then, we present our experiments which were conducted to evaluate QS agents' performance in solving lunar-lander problem compared to RL agents. 

\begin{figure}[ht]
	\vskip 0.2in
	\begin{center}
		\centerline{\includegraphics[width=\columnwidth]{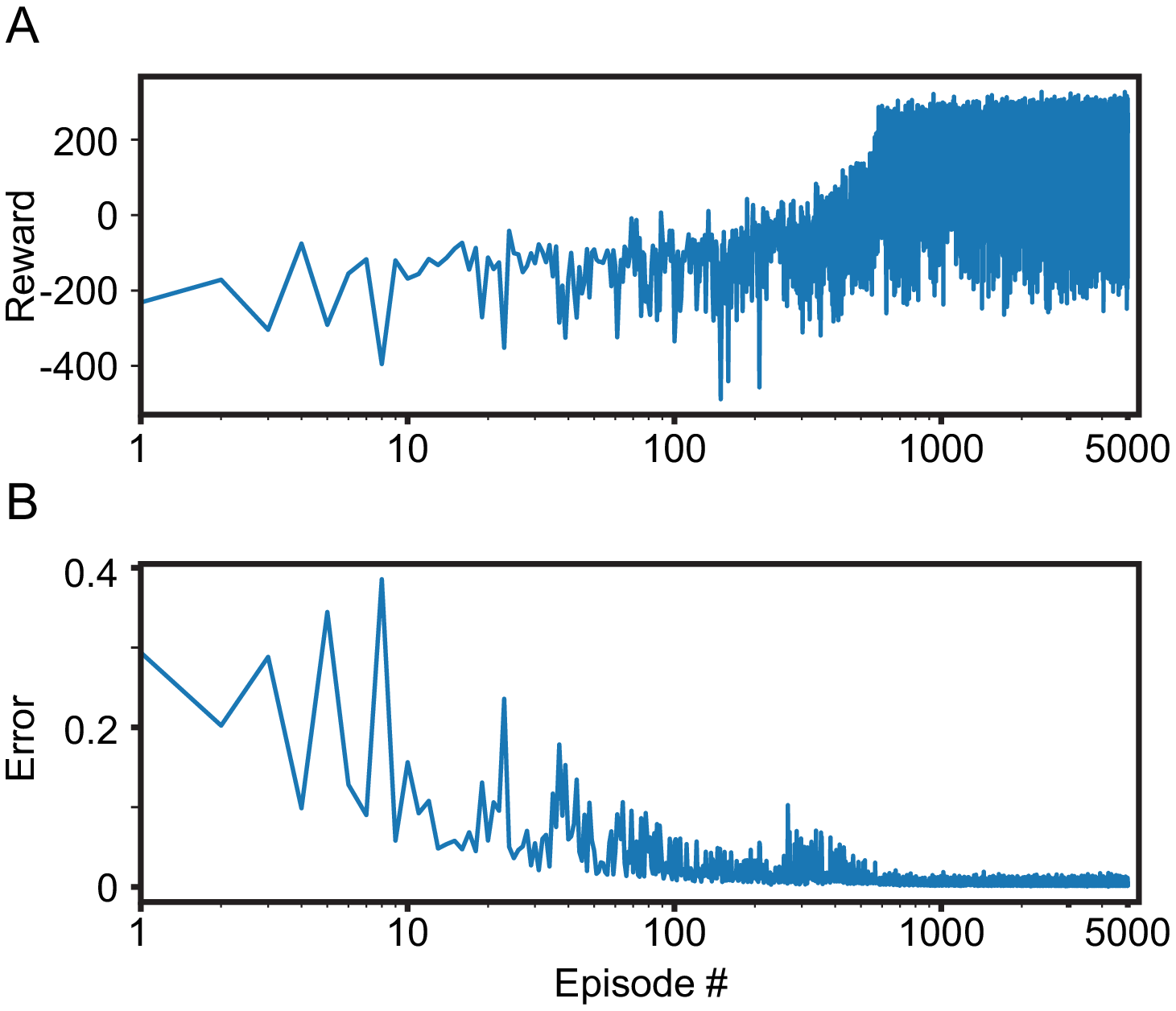}}
		\caption{Training of RL agent and Env network. (A), the total reward in each episode. (B), the error observed during training of the Env network. }
		\label{Fig2}
	\end{center}
	\vskip -0.2in
\end{figure}

\subsection{Training and operating rules of Quasi-Symbolic (QS) agents}

Unlike RL agents, QS agents do not work alone. Instead, their operations depend on RL agents in both learning and inference modes; that is, RL and QS agents are complementary with each other. The matching and value networks in QS agents are updated by using RL agents' behaviors during training. Specifically, in each time step in the training period, the previous state $S$ and the current state $S'$ are used to generate the state transition vectors $\Delta S=S'-S$, and this transition vector is fed into the matching network as inputs. The matching network first determines the novelty of the current transition vector; this novelty detection is done by inspecting synaptic inputs to matching nodes (see section \ref{network structure} and Eq. \ref{eq1} for details). When a novel transition vector is introduced, a new output node is added to the matching and value networks (Fig. \ref{Fig1}A), and the connection between matching and value networks (one-to-one connection $w_k$ between newly added nodes, $m_k$ and $v_k$) is established (section \ref{network structure}). The strength of the connection $w_k$ is determined by the reward given to the RL agent after the transition from $S$ to $S'$. When the earlier input is introduced again, the maximally activated matching node is identified, and the connection between the identified matching node and the value node, which is exclusive and one-to-one (Fig. \ref{Fig1}A), is updated by adding the reward obtained to the previous strength (Eq. \ref{eq2}). It should be noted that only one matching node is allowed to be active and used to assess the value of the transition. In brief, the matching network memorizes transition vectors observed during training, and the value network stores the amount of rewards induced by the observed transitions. 

In the inference mode, in which QS agents choose the best action, QS agents utilize both the trained RL agents and Env networks to make action plans, whose lengths are variable. In doing so, QS agents identify the most valuable transition vectors $\Delta S_k$ by inspecting connections' strength between matching and value networks. In this study, we identified synaptic weights higher than $\theta_{hub}$ defined in Eq. \ref{eq3}.

\begin{equation}\label{eq3}
\theta_{hub}=<w_k>+\alpha \times \sigma(w_k)
\end{equation} 

, where $<w_k>$ and $\sigma(w_k)$ are mean and standard deviations of synaptic weights, and $\alpha $ is the scale constant which determines the magnitude of $\theta_{hub}$. The default value of $\alpha$ is $0.1$. Due to the one-to-one connections between matching and value networks, the synaptic weights allow us to identify the most valuable transitions observed during the training period, which are referred to as hub states hereafter. 

\begin{figure}[ht]
	\vskip 0.2in
	\begin{center}
		\centerline{\includegraphics[width=0.55\columnwidth]{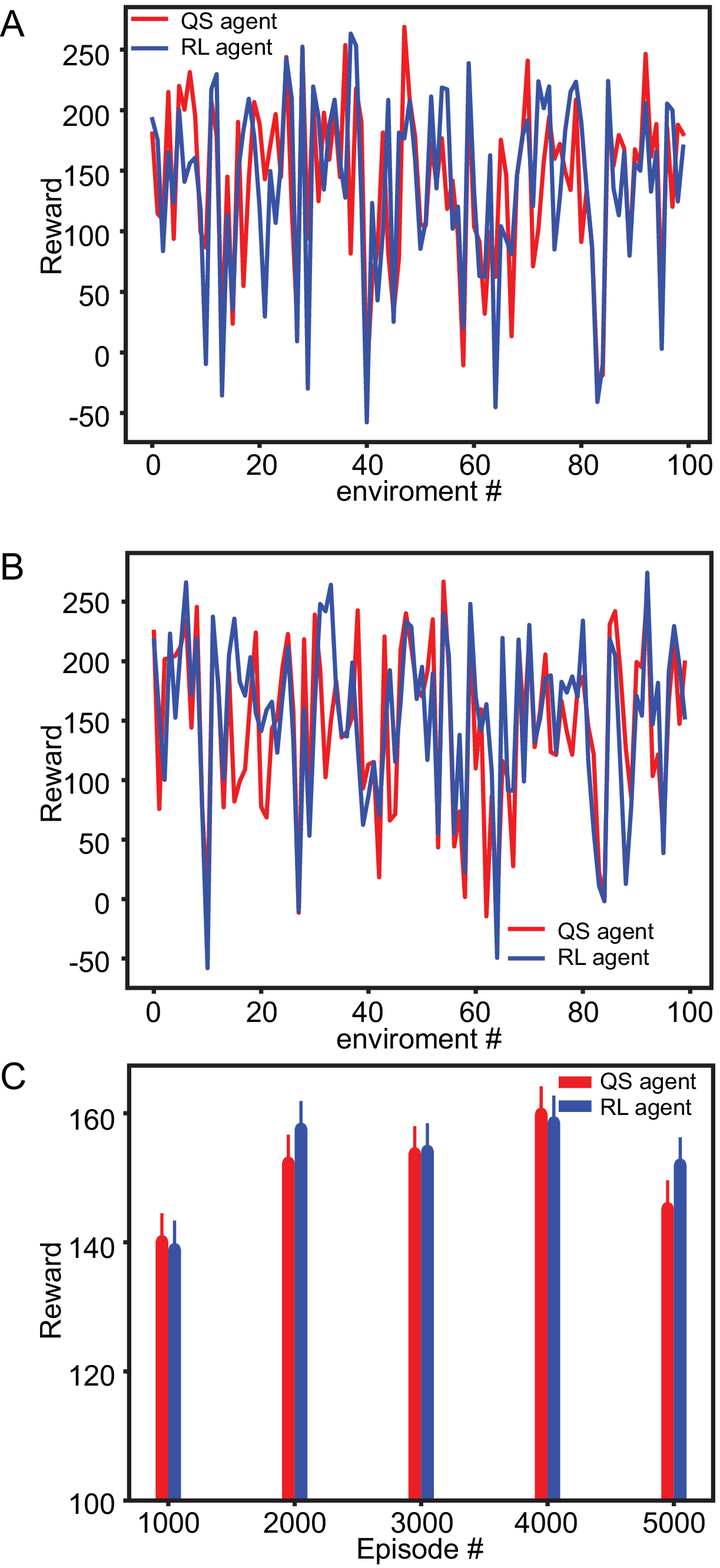}}
		\caption{Comparison between QS and RL agents' performances. (A), the average reward of QS and RL agents in 100 environments, after the RL agent is trained in 1000 episodes. 10 independent QS and RL agents are tested for the same environment, and the rewards averaged over 10 agents are shown. The red and blue lines represent QS and RL agents, respectively. (B), the same as (A), but the RL agent is trained during 5000 episodes. (C), the average reward of QS and RL agents depending on the length of RL agents training. The mean values of the rewards are estimated using 10 agents tested in 100 environments. That is, the mean values illustrated are rewards averaged over 1000 individual experiments. The error bars are standard errors from the same 1000 experiments.}
		\label{Fig3}
	\end{center}
	\vskip -0.2in
\end{figure}

After identifying the hub states, QS agents search for an action plan to reach one of the hub states. In doing so, QS agents utilize the Env network and the trained RL agent to predict future states (and thus transitions as well) to make an action plan. At each state, RL agent provides a possible action, and Env network returns the next state in response to the suggested action. Employing them recursively (Fig. \ref{Fig1}B), QS agents can predict the future states. During this planning, at each time step, QS agents examine whether a transition vector is one of the hub states (i.e., the most valuable transitions) or not. If QS agents do expect to reach one of the hub states, they stop planning and execute the current plan. The maximum length of the action plan is 10 time-step, unless stated otherwise. If no hub state cannot be reached within the predefined maximal time-step, QS agents start over and make a new plan. For each state, QS agents are allowed to make a total of 5 different plans. If they cannot find a path to the hub states in all five plans, they inspect the values of the first transition in the five plans and choose the best immediate action according to the reward estimated by the value network. In this case, rather than taking a sequence of actions, QS agents execute a single action (i.e., the first action in the plan) only. 

\subsection{QS agents' performance compared to RL agents}

To evaluate QS agents' performance, we compared their performance to that of RL agents by using the lunar-lander benchmark task included in the OpenAI gym environment \cite{Brockman2016}. In this study, we constructed RL agents (actor-critic model), QS agents and Env network using Pytorch, an open-source machine learning library \cite{Paszke2017}; their schematics are illustrated in Fig. \ref{Fig1}. We trained RL, QS agents and Env network during 5000 episodes; see section \ref{network structure} for training details. Figures \ref{Fig2}A and B show the total amount of rewards given to the RL agent and the error function of the Env network during 5000 episodes. As shown in the figure, while the reward in each trial fluctuates from one trial to another, the amount of rewards, on average, increases rapidly until 1000 episodes. After 1000 episodes, the speed of improvement is reduced. Similarly, the error of Env network is reduced most rapidly in the first 1000 episodes, and then the speed of error-reduction slows down. 

At every 1000 episode, we froze the learning and tested both QS and RL agents using the same 100 environments of lunar lander; that is, the environments are instantiated with the same random seeds. The RL agent in this study uses stochastic policy to choose actions (eq. \ref{eq_grad}), and thus its behaviors depend on the random seed forwarded to the Pytorch. Moreover, QS agents' behaviors are also stochastic, as they rely on RL agents for their decisions. To avoid potential biases based on their stochastic behaviors, we constructed 10 independent QS and RL agents by forwarding distinct random seeds to pytorch and calculated the average reward for both agents. Figures \ref{Fig3}A and B show the average amount of QS and RL agents for all 100 instantiations of environment. $x$-axis represents the identity of environment, and $y$-axis represents the reward averaged over 10 independently constructed agents. Rewards are estimated after training the RL agent in 1000 episodes (Fig. \ref{Fig3}A) and 5000 episodes (Fig. \ref{Fig3}B). As shown in the figure, the performance of QS agents (shown in red) with 10 time-step action plan are comparable to that of RL agents (shown in blue). Figure \ref{Fig3}C shows the average reward calculated using 10 independent agents in 100 environments, after training the RL agent in 1000, 2000, 3000, 4000 and 5000 episodes, respectively; that is, the mean values and standard errors are estimated from 1000 individual experiments (10 independent agents $\times$ 100 environments). 

\begin{figure}[ht]
	\vskip 0.2in
	\begin{center}
		\centerline{\includegraphics[width=0.8\columnwidth]{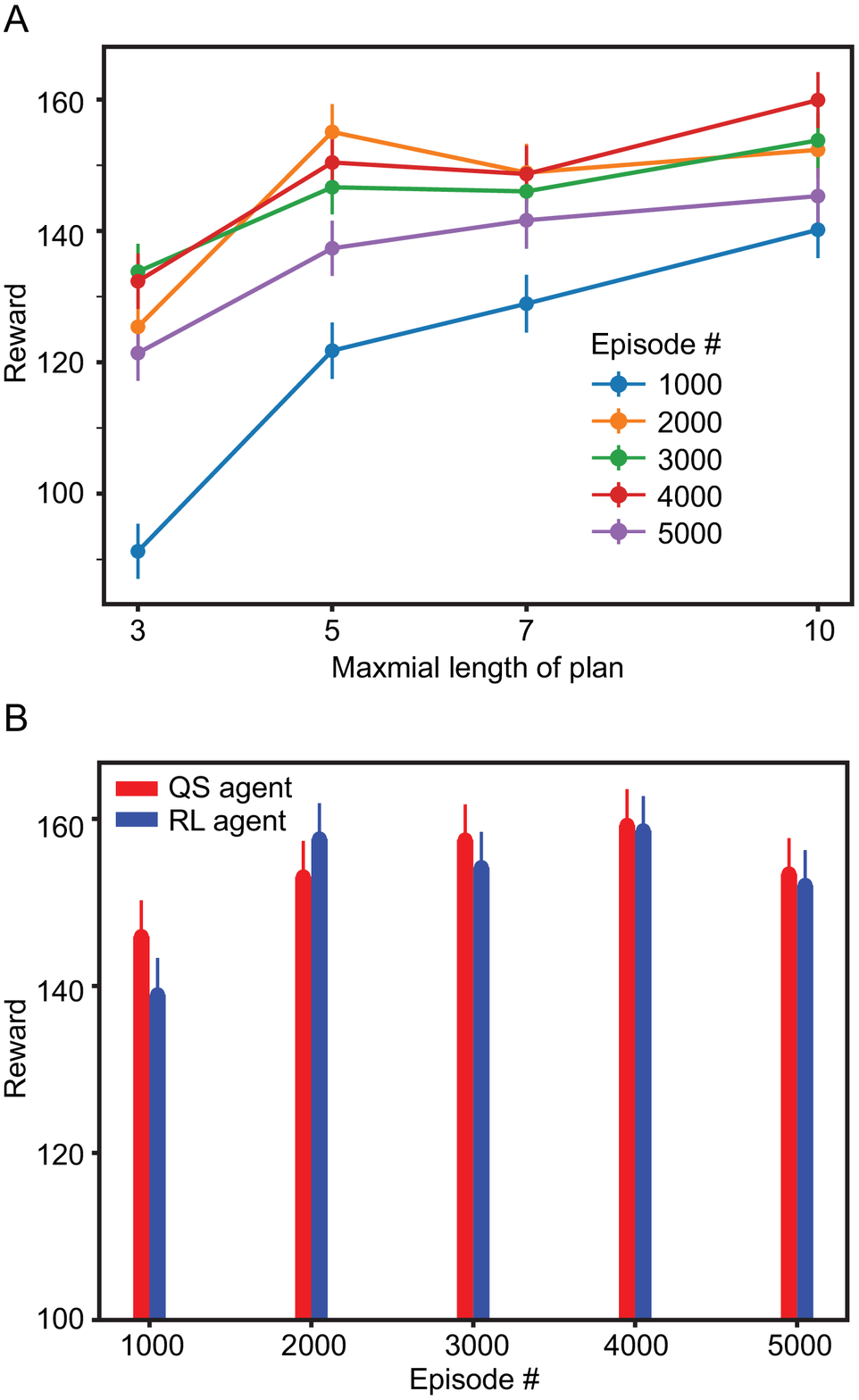}}
		\caption{Parameters' effects on the QS agents' performance. (A), the average rewards depending on the maximal length of action plans. The mean values and standard errors of the rewards are estimated using 10 agents tested in 100 environments. We estimated them with RL agents trained during 1000, 2000, 3000, 4000 and 5000 episodes. The color codes represent the length of RL agents training. (B), the average rewards with a bigger set of hub states. The threshold value for the hub states is lowered ($\alpha$ in Eq. \ref{eq3}: $0.1 \rightarrow 0.05$) to increase the number of states.}
		\label{Fig4}
	\end{center}
	\vskip -0.2in
\end{figure}

\begin{figure}[ht]
	\vskip 0.2in
	\begin{center}
		\centerline{\includegraphics[width=0.8\columnwidth]{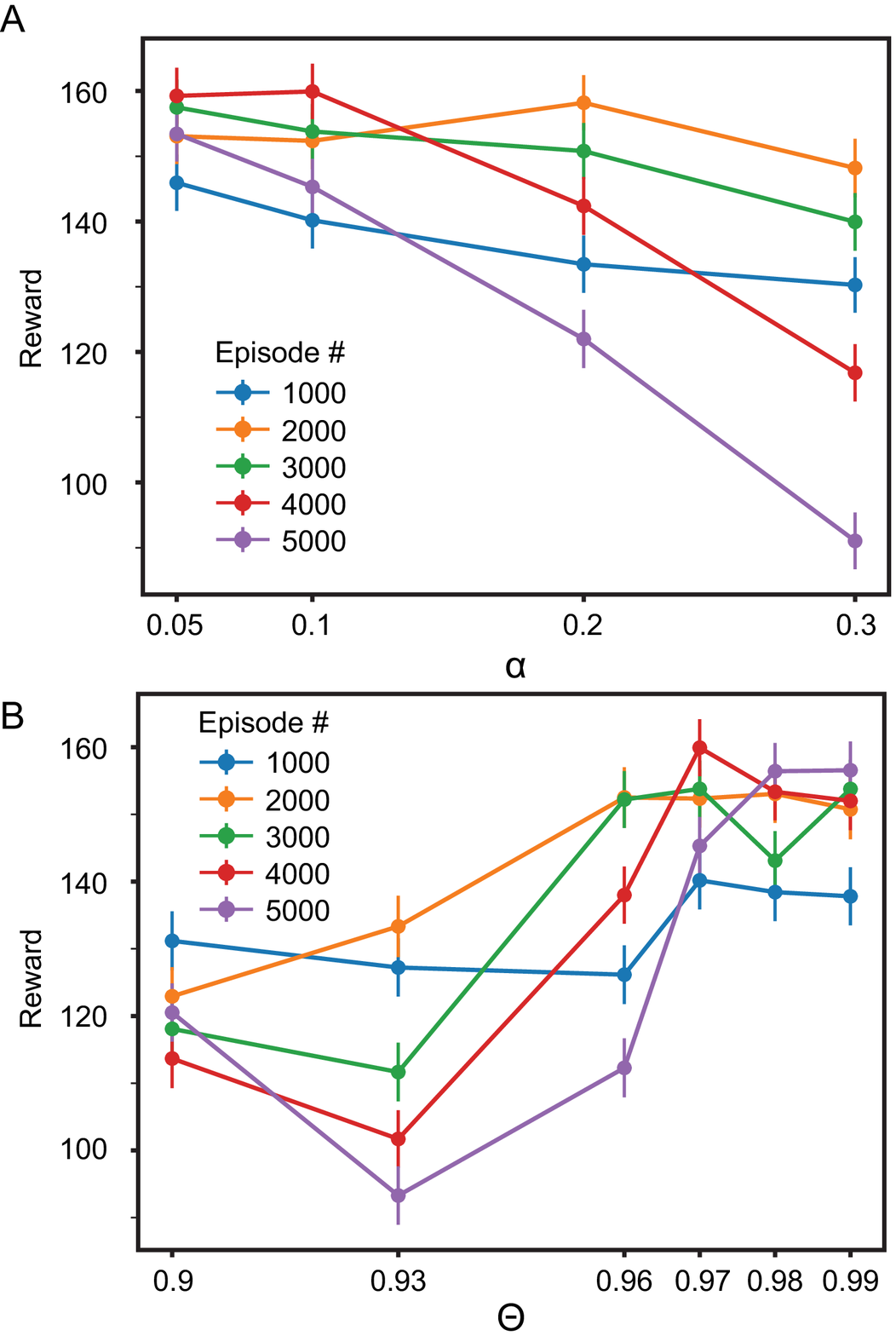}}
		\caption {Parameters' effects on the QS agents' performance. (A), the average rewards depending on the threshold value for the hub states (see Eq. \ref{eq3}). (B), the average rewards depending on the threshold value $\theta $ for novel input detection (section \ref{network structure}).}
		\label{Fig5}
	\end{center}
	\vskip -0.2in
\end{figure}

Then, we varied the parameters to examine how they affect QS agents' performance. First, we tested the effects of action plans' lengths. As shown in Fig. \ref{Fig4}A, QS agents' performance improves, as the maximal length of action plans increases. Second, we increased the number of hub states by lowering $\theta_{hub}$. Specifically, we set $\alpha=0.05$ and estimated the amount of rewards. As shown in Fig. \ref{Fig4}B, QS agents' performance improved. We also increased $\theta_{hub}$ to further examine the effects of the number of hub states on QS agents' performance and found that QS performance is negatively correlated with $\theta_{hub}$ (Fig. \ref{Fig5}A). Finally, we perturbed the threshold value $\theta $ for the novelty detection (used in matching networks) to see how it affects QS agents' performance. As shown in Fig. \ref{Fig5}B, QS agents obtained less rewards. This may be explained by the fact that the accuracy of predictions on QS agents' future states decreases, as the threshold $\theta $ becomes lower,

\section{Discussion}

In this study, we propose QS agents to develop transparent RL agents. The newly proposed QS agents have two operating units, matching and value networks. With these two units, QS agents evaluate actions suggested by RL agents and choose the most probable choice. To select the most probable action, QS agents search for a path to reach one of the hub states by utilizing the Env network which models the environment. Our results (Fig. \ref{Fig4}) suggest that this future plan ensures QS agents’ good performance. 

Then, do QS agents have transparent decision processes? QS agents’ decisions are transparent for two reasons. First, the two operating units of QS agents have simple structures, which can be analyzed easily. The value network simply accumulates rewards given after the transition of states into synaptic weights, and it returns these stored values depending on inputs ($\Delta S$). That is, the value network is, in principle, equivalent to conventional memory units. The matching network identifies the old input ($\Delta S$), which is the closest to the current input by using the cosine similarity (eq. \ref{eq1}). In addition, as only a single matching node is allowed to be active, it is clear that the matching network's operation can be easily analyzed. Second, QS agents rely on a simple set of rules to select the best actions based on suggestions made by RL. With both simple inner mechanisms and operating rules, QS agents have transparent decision-making process. 

Moreover, it should be noted that the output nodes of the matching network are working independently from each other. That is, the matching nodes can be removed and added without interfering with other matching nodes' operations. This property makes manual modification of QS agents' actions possible. If a new piece of evidence finds a particular state-transition to be unacceptable, it can be removed. On the other hand, if some state-transitions $\Delta S$, which have not been previously observed, are considered valuable, they can be added to QS agents. Similarly, the synaptic weights of value networks and hub states can also be changed, if necessary. Therefore, QS agents can be continuously and incrementally improved (or repaired) to avoid mistakes.    

\subsection{Potential variations of QS agents}

To address the possibility of deriving a comprehensible secondary agent from RL agents, we sought generic algorithms to be applied to general RL problems, but QS agents and their operations can be customized in domain-specific ways. Below, we list a few potential variations of our generic QS agents. 

First, current QS agents evaluate individual transitions $\Delta S$ using the rewards RL agents obtain immediately after completing the transition. However, the actual values of transitions can be estimated differently. For instance, instead of using the immediately obtained reward, we can use the total rewards from the transition to the end of an episode for evaluating state transitions. 

Second, in this study, QS agents treat all state variables equally and rely on a single-state vector $\Delta S$. However, state variables do not have equivalent values in agents' behaviors. For example, the coordinates and velocities included in the state vectors of lunar-lander do have different importance in terms of agents dynamics. If we deal with coordinates and velocities distinctively, QS agents may evaluate the state-transitions more effectively. In this case, there will be two state vectors; $S \rightarrow S_{\dot{x}}$, $S_{x}$. Similarly, a single state-transition vector can be split into multiple ones; $S \rightarrow S_1, S_2, ..., S_n$. In doing so, QS agents need to utilize multiple matching and value networks, and the actual value can be estimated with a linear summation of outputs of multiple value nodes.  

Third, the state transition vectors $\Delta S$ can be coupled with state vectors $S$ to estimate the values of actions more precisely. For instance, an agent's horizontal move can be either bad or good depending on the current state. If both state and transition vectors are used, QS agent may have better estimation of agents' actions.

\subsection{Implications for the brain' complementary system}

Prefrontal cortex (PFC) has been thought to be a hub for high-level cognitive functions such as decision-making, learning and working memory \cite{Lara2015, Miller2001, Tanji2008, Wang2012}. However, PFC does not work alone and is known to be connected to other brain areas. The two main areas that have strong interactions with PFC are hippocampus \cite{Jin2015, Peyrache2011} and anterior cingulate cortex (ACC) \cite{Gehring2000, Paus2001}. Notably, complementary learning system theory suggests central roles of the interplay between PFC and hippocampus in our ability to learn continuously \cite{OReilly2014}. Then, why does PFC need to interact with ACC which has been postulated to be associated with multiple functions such as error likelihood \cite{Brown2005} and prediction of expected outcomes \cite{Vassena2014}? We note that QS agents can predict the future outcomes, consistent with one of ACC’s hypothetical functions. Based on our results that QS agents can evaluate RL agents' decisions, we propose that one of ACC functions is to evaluate possible actions suggested by PFC and choose the best one depending on context.

\bibliographystyle{unsrt}  


\end{document}